\documentclass{article}

    \usepackage[preprint]{neurips_2022}

\usepackage[utf8]{inputenc} 
\usepackage[T1]{fontenc}    
\usepackage{url}            
\usepackage{booktabs}       
\usepackage{amsfonts}       
\usepackage{nicefrac}       
\usepackage{microtype}      
\usepackage{xcolor}         
\usepackage{times}
\usepackage{epsfig}
\usepackage{graphicx}
\usepackage{amsmath}
\usepackage{amssymb}
\usepackage{multirow}
\usepackage{adjustbox}
\usepackage[font=small]{caption}
\usepackage{enumitem}
\usepackage{stmaryrd}
\usepackage{pifont}
\usepackage{xspace}
\usepackage{subcaption}
\usepackage{makecell}
\usepackage[pagebackref=true,breaklinks=true,colorlinks,bookmarks=false]{hyperref}
\usepackage{listings}

\makeatletter
\DeclareRobustCommand\onedot{\futurelet\@let@token\@onedot}
\def\@onedot{\ifx\@let@token.\else.\null\fi\xspace}

 \def\vs{\emph{vs}\onedot}
 
\def\etal{\emph{et al}\onedot}
\makeatother

\newcommand{\system}{PS-NeRV\xspace}
\title{PS-NeRV: Patch-wise Stylized Neural Representations for Videos}

\author{%
  Yunpeng Bai$^{1}$, Chao Dong$^{2,3}$ , Cairong Wang$^{1}$ \\[0.5em]
  $^{1}$Shenzhen International Graduate School, Tsinghua University,\\[0.3em]  $^{2}$Shenzhen Institutes of Advanced Technology, Chinese Academy of Sciences, \\[0.3em] $^{3}$Shanghai AI Laboratory, China \\[0.3em]
}

\begin{document}

\maketitle

\begin{abstract}
We study how to represent a video with implicit neural representations (INRs). Classical INRs methods generally utilize MLPs to map input coordinates to output pixels. While some recent works have tried to directly reconstruct the whole image with CNNs. However, we argue that both the above pixel-wise and image-wise strategies are not favorable to video data. Instead, we propose a patch-wise solution, PS-NeRV, which represents videos as a function of patches and the corresponding patch coordinate. It naturally inherits the advantages of image-wise methods, and achieves excellent reconstruction performance with fast decoding speed. The whole method includes conventional modules, like positional embedding, MLPs and CNNs, while also introduces AdaIN to enhance intermediate features. These simple yet essential changes could help the network easily fit high-frequency details. Extensive experiments have demonstrated its effectiveness in several video-related tasks, such as video compression and video inpainting. 
\end{abstract}

\section{Introduction}
\label{section1}

With the fast development of streaming media, numerous video data have been widely filled in our daily life. However, the large file sizes, especially for high-resolution (1080p-4k) videos, are becoming heavy burdens to storage and transmission. Traditional video representations, which explicitly represent videos as frame sequences, are not efficient enough to meet this challenge.
Recently, Implicit Neural Representations \cite{DBLP:journals/tog/MartelLLCMW21,DBLP:conf/eccv/MildenhallSTBRN20,DBLP:conf/cvpr/TancikMWSSBN21,DBLP:conf/nips/TancikSMFRSRBN20} (INRs) have gained increasing attention as a novel and effective representation method that is able to produce high-fidelity results for various data types, such as images \cite{DBLP:conf/nips/SitzmannMBLW20}, 3D shapes \cite{DBLP:conf/cvpr/ParkFSNL19}, and scenes \cite{DBLP:conf/eccv/MildenhallSTBRN20}. Specially, using neural network to represent video implicitly has also shown its great potential.

For the implicit video representation, pixel-wise methods (like SIREN \cite{DBLP:conf/nips/SitzmannMBLW20}) output the RGB value for each pixel according to the spatio-temporal coordinates (x,y,t). 
In contrast, NeRV \cite{DBLP:conf/nips/ChenHWRLS21} is proposed as an image-wise representation method, which represents a video as a function of time. It maps each timestamp $t$ to an entire frame, and shows superior efficiency to pixel-wise representation methods.
However, neither pixel-wise nor image-wise representation is the most suitable strategy for video data. They could increase the network burden in different ways, resulting in unsatisfactory reconstruction results. Specifically, the pixel-wise method requires a large number of samples for each frame, which is inefficient for both encoding and decoding. The image-wise method could be struggle to represent complex signals, such as high-resolution videos. It requires a much larger network to overfit the content and details of the whole frame, bringing additional computation cost. Therefore, we need a more appropriate and effective way to implicitly represent video data.

Inspired by the similarity of local adjacent pixels (widely exists in real-world signals \cite{roweis2000nonlinear}), we propose a patch-wise implicit representation method of videos. As the features in adjacent patches have great similarities, they can be easily represented by a single network. In our implementation, each frame of the video is divided into split patches, so that each patch has a corresponding spatial coordinate. We take the patch coordinate and timestamp as the input of the network, then the network outputs the corresponding image patch by convolutional network. Our method not only enjoys the fast encoding and decoding speed as the image-wise method, but also reconstructs vivid high-frequency details. Note that patch-wise representation is not a trade-off between pixel-wise and image-wise method, but a more suitable solution for video data.

Another point worth noting is that the normalization layer widely used in convolution neural networks could reduce the fitting ability of the network, which is also found in NeRV \cite{DBLP:conf/nips/ChenHWRLS21}. As an alternative, can we improve the capability of fitting by aligning the mean and variance of features with the target frame?
Based on the above consideration, we further introduce the Adaptive Instance Normalization layer (AdaIN) \cite{DBLP:conf/iccv/HuangB17} to modulate features. The mean and variance of the frame features can be directly learned from the input coordinate through an MLP network. This simple strategy, which is similar to StyleGAN \cite{DBLP:conf/cvpr/KarrasLA19}, can significantly improve the fitting ability of the network, especially for details. Finally, we name our whole method as Patch-wise Stylized Neural Representations for Videos (PS-NeRV).

We also explore some applications of our method, such as video compression and video 
inpainting tasks. Compared with NeRV, our method shows better compression potential.
When the input video is masked, our method can generate high-quality inpainting output and even outperforms state-of-the-art video inpainting methods.

To summarize, the main contributions of this work are as follows:

\begin{itemize}
\item We design a novel video implicit representation method, which represents the video as image patches, and verify the effectiveness of the proposed method in representing the details. 

\item We find that introducing the AdaIN layer to the ConvNets-based INR can significantly improve the fitting effect of the model.

\item Our method shows excellent performance in several video-related applications, including video compression and video inpainting.
\end{itemize}

\begin{figure}
    \centering
    \includegraphics[width=.98\linewidth]{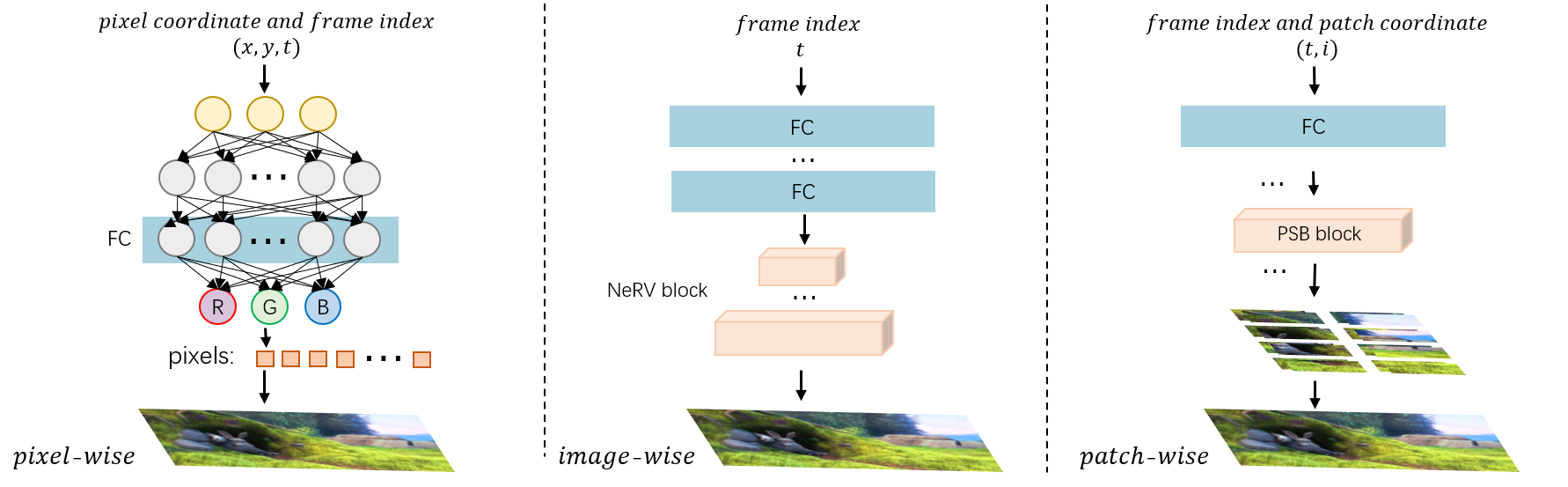}
    \vspace{-0.05in}
    \caption{Comparison of three different INR methods.}
    \label{fig:com_inrs}
    \vspace{-0.1in}
\end{figure}

\begin{figure}
    \centering
    \includegraphics[width=.98\linewidth]{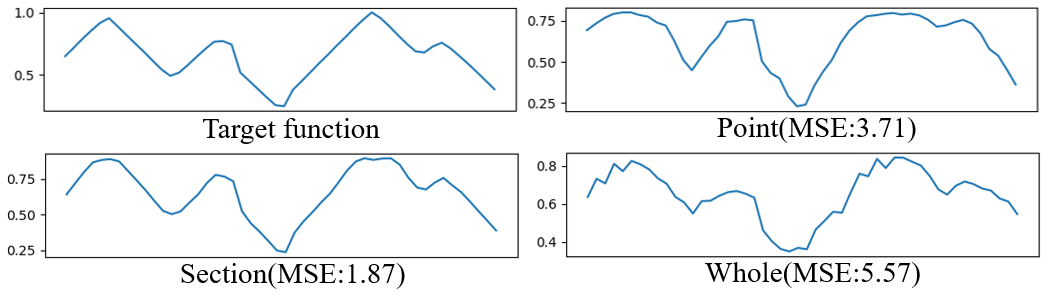}
    \vspace{-0.05in}
    \caption{We use three MLP networks of the same size to fit a function. Their outputs are the values of each point, the values of each section, and the values of the whole curve, respectively. We trained the three networks for the same time, and the section-wise fitting way is the most effective.}
    \label{fig:motivation}
    \vspace{-0.1in}
\end{figure}


\section{Related Work}
\label{section2}
\subsection{Implicit Neural Representation} 

The Implicit Neural Representation (INR) approach has recently gained increasing attention and is used in several tasks, such as images \cite{DBLP:conf/nips/SitzmannMBLW20}, 3D shapes \cite{DBLP:conf/cvpr/ParkFSNL19}, and novel view synthesis \cite{DBLP:conf/eccv/MildenhallSTBRN20}. In some later works, it is found that using periodic activation function \cite{DBLP:conf/nips/SitzmannMBLW20} and the so-called Fourier feature \cite{DBLP:conf/nips/TancikSMFRSRBN20} to encode pixel (or voxel) coordinates can effectively reconstruct fine details of signals. However, INR methods are basically coordinate-based methods. It means that for data with more pixels, such as video data, it is necessary to a use neural network to build mapping from coordinates to RGB value. Numerous samples will also lead to the low efficiency of training and testing, which makes the INR methods unable to be applied in some practical scenarios. NeRV \cite{DBLP:conf/nips/ChenHWRLS21} explores a 
image-wise representation, which builds a mapping from timestamp $t$ to an entire frame, and shows its excellent efficiency compared to pixel-wise representation. However, NeRV \cite{DBLP:conf/nips/ChenHWRLS21} increases the burden of the network to represent the complex signal of a whole frame, and the architecture design of its neural network is also relatively rough. In contrast, we use neural network to represent simpler signals in the patch, which ensures efficiency and further improves accuracy.

\begin{figure}
    \centering
    \includegraphics[width=.98\linewidth]{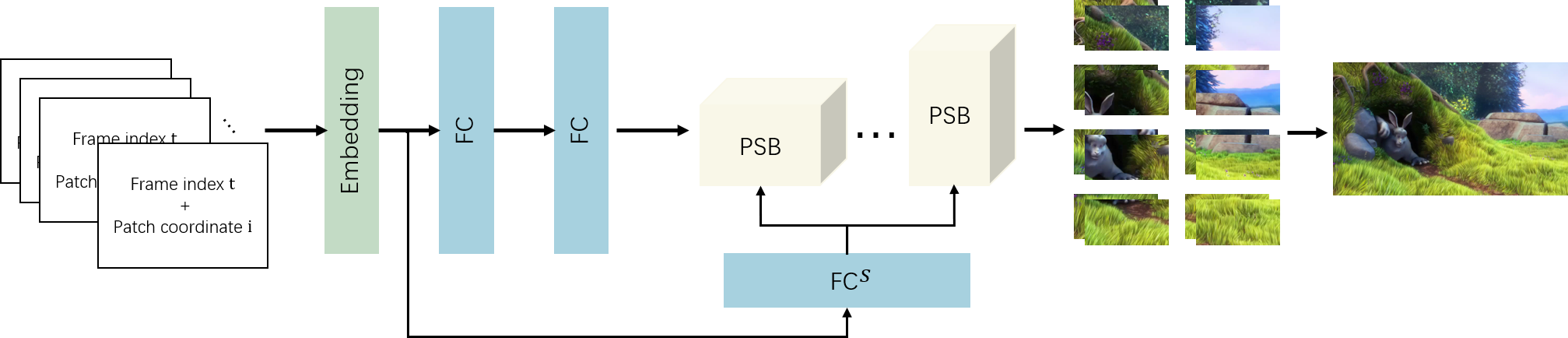}
    \vspace{-0.05in}
    \caption{The input of our patch-wise architecture is time and patch coordinates, the output is image patch, and the input is also used to modulate the features of the later layer. Our method takes into account both efficiency and accuracy.}
    \label{fig:arch1}
    \vspace{-0.1in}
\end{figure}


\subsection{Video Compression}

Video compression is a fundamental task that has been studied for a long time. A lot of traditional video compression algorithms have been proposed and achieved great success in the past decades, such as H.264~\cite{wiegand2003overview}, MPEG~\cite{le1991mpeg} and HEVC~\cite{sullivan2012overview}. Recently, several deep-learning-based methods have attempted to utilize neural networks to improve video compression. DVC~\cite{lu2019dvc} utilizes neural networks to replace all the key components in H.264, and has achieved compression rates that are comparable to or slightly better than traditional video compression algorithms. Later, Hu et al. \cite{DBLP:conf/cvpr/HuL021} propose a feature-space video coding network by performing all major operations (i.e.,motion estimation, motion compression, motion compensation and residual compression) in the feature space. SRVC \cite{DBLP:conf/iccv/ShirkoohiSA21} adds another model stream to the traditional compression method, and decodes the video by passing the decompressed low-resolution video frames through the (time-varying) super-resolution model to reconstruct high-resolution video frames. Li et al. \cite{DBLP:conf/nips/LiLL21} propose to use context in feature domain to help the encoding and decoding. However, the overall pipeline of traditional compression still limits the capabilities of these methods. On the contrary, NeRV \cite{DBLP:conf/nips/ChenHWRLS21} uses the INR method to convert video compression task into a model compression problem, and shows great potential. In most scenes, one video is encoded once but will need to be decoded many times. Therefore, the INR method such as NeRV \cite{DBLP:conf/nips/ChenHWRLS21} shows great advantages because of its high decoding efficiency. On the other hand, compared with other video compression methods that need to be decoded in a sequential manner after the reconstruction of the respective key frames, INR methods make parallel decoding very easy.

\subsection{Local Implicit Functions}

Recently, some coordinate-based methods also use local features to represent complex signals with higher accuracy, such as images \cite{DBLP:conf/cvpr/ChenL021}, shapes \cite{DBLP:conf/cvpr/MeschederONNG19} and radiance fields \cite{DBLP:conf/nips/LiuGLCT20, DBLP:journals/tog/MartelLLCMW21}. These methods first decompose the target domain into an explicit grid, and estimate a local continuous representation on each grid. Then a decoder will output the value of each coordinate according to this representation. Since each grid cell needs to store a latent code or feature vector that the local network is conditioned, these methods will be less memory efficient. Instead, we map the grid coordinates directly to the whole grid output value, which is effective and more efficient. Similar to our approach, COCO-GAN \cite{DBLP:conf/iccv/LinCCJ0C19} generates images by parts based on their spatial coordinates as the condition. However, their method is difficult to ensure the correct spatial connection between each generated part and reduces the flexibility. Our method is to fit complex signals in a patch-wise way.














\begin{figure}
    \centering
    \includegraphics[width=.98\linewidth]{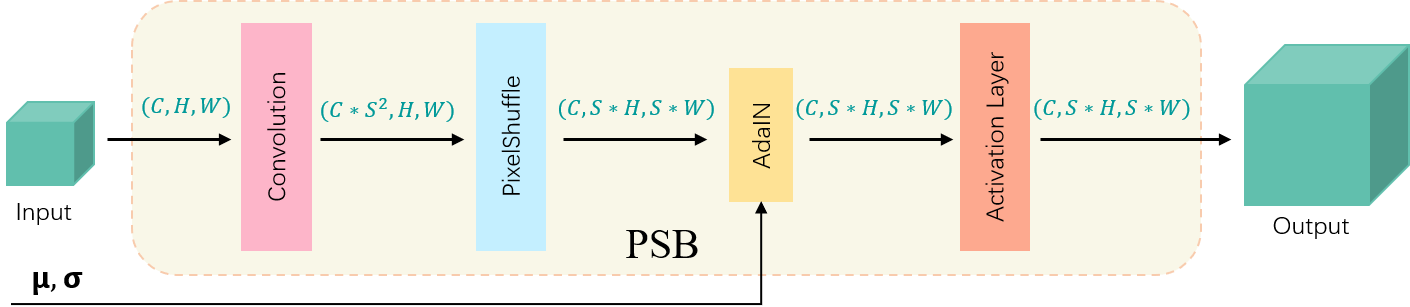}
    \vspace{-0.05in}
    \caption{\textbf{ Patch-wise stylized block }architecture, S is the upscale factor.}
    \label{fig:arch2}
    \vspace{-0.1in}
\end{figure}
\section{Proposed Method}
\label{section3}
\subsection{Motivation}

There are local similarities in various kinds of real-world signals, which make the structure of local features easy to represent, like Locally Linear Embedding (LLE) \cite{roweis2000nonlinear}. Therefore, we conjure that representing signals with local patterns is the most effective way. To facilitate understanding, we conduct a toy experiment as shown in Figure \ref{fig:motivation}. Specifically, we try to fit a non-linear function (curve) with three different ways, namely point-wise, section-wise and whole-wise strategies. The same MLP is used to train these three methods. Implementation details can be found in the supplementary file.  From the results in Figure \ref{fig:motivation}, it is clearly observed that the section-wise method achieves the best fitting performance. In comparison, the pixel-wise method cannot reconstruct the correct trend. While the whole-wise strategy tends to add some undesired high-frequency details, which can be considered as artifacts. This partially demonstrates our primal assumption. The point-wise, section-wise and whole-wise can be analogs to pixel-wise, patch-wise and image-wise, respectively. The above phenomenon could inspire us to use patch-wise strategy as a more effective way of implicit representation. Next we will describe our method PS-NeRV in detail.

\subsection{Represent videos as images patches}

For each video $V = \{v^t\}^T_{t=1} \in \mathbb{R}^{T\times H\times W \times 3}$,  we split each frame into $N*N$ patches and get $\{v^t_{p}\}^{N^2}_{p=1} \in \mathbb{R}^{N^2 \times H/N\times W/N \times 3}$.
Then, these split patches will be implicitly represented by a function $f_\theta: \mathbb{R} \rightarrow \mathbb{R}^{H/N\times W/N \times 3}$, which parameterized with a deep neural network $\theta$,  $v^t_{p_i} = f_\theta(t,i)$.
The function has two inputs:  a frame index $t$ with a patch coordinate $i$, and the output is the corresponding patch image $v^t_{p_i}\in \mathbb{R}^{H/N\times W/N \times 3}$. 
Therefore, we construct a mapping from spatio-temporal coordinates to image patches through this neural network $f_\theta$.
After obtaining all the patches, we can directly splice them into an entire frame $v^t\in \mathbb{R}^{ H\times W \times 3} $.

\subsection{Time-Coordinate Embedding}

When the coordinates are used as the input of the neural network, it has been found ~\cite{rahaman2019spectral,mildenhall2020nerf} that mapping them to a high embedding space can effectively improve the fitting effect of the network.
In addition to the patch coordinate $i$, there is another input -- timestamp $t$. We encode both two inputs into embedding using Positional Encoding ~\cite{mildenhall2020nerf,vaswani2017attention,tancik2020fourier}function:


\begin{equation}
    \begin{split}
    \Gamma(t,i) = (\sin\left(b^0\pi t\right), \cos\left(b^0\pi t\right), \dots, \sin\left(b^{l-1}\pi t\right), \cos\left(b^{l-1}\pi t\right);\\
    \sin\left(b^0\pi i\right), \cos\left(b^0\pi i\right), \dots, \sin\left(b^{l-1}\pi i\right), \cos\left(b^{l-1}\pi i\right)),
    \end{split}
    \label{equa:input-embed}
\end{equation}

where $b$ and $l$ are hyper-parameters of the networks. 
According to the length of the video and the number of patches, timestamp $t$ and coordinate $i$ are normalized between $(0,1]$. Then, their embeddings will be concatenated together as the input of the network.

\subsection{Network Architecture}

The input of the time-coordinate embedding is sent to the subsequent layers of MLPs to get a suitable size for the later block. The latter patch-wise stylized block (PSB) then gradually recovers it to an image patch.
Our PSB consists of convolution and up-sampling layers. The AdaIN module is followed after each up-sampling layer.
With the joint effect of the two conditions -- time and coordinate, the network outputs patches instead of whole image. Such a practice greatly reduces the burden of the network. Experiments also show that our model is easier to fit the details than image-wise method, i.e., NeRV.




\subsection{Time-Coordinate Stylization}

The training process is to over-fit a video.
NeRV has found that the normalization layer widely used in convolution neural networks could reduce the fitting ability of ConvNets-based INR. On the contrary, we find that aligning the mean and variance of features with the target frame can speed up the fitting process and obtain higher quality results.

We use Adaptive Instance Normalization (AdaIN) \cite{DBLP:conf/iccv/HuangB17} to modulate the features of the later convolution layers according to the time-coordinate condition:

\begin{equation}
  \mathbf{AdaIN}(x_i,\sigma^s_{i},\mu^s_{i})=\sigma^s_{i}\frac{x_i-\mu(x_i)}{\sigma(x_i)}+\mu^s_{i}\ ,
  \label{equ:adain}
\end{equation}

where $\mu(x_i)$ and $\sigma(x_i)$ represent the $i^{th}$ feature map's mean and variance, respectively.
We use another MLP network to learn $\sigma^s$ and $\mu^s$ required by the later AdaIN:

\begin{equation}
    \begin{aligned}
        (\sigma^s,\mu^s) = \mathbf{MLP}_\mathcal{S}(\Gamma(t,i)).
    \end{aligned}
\end{equation}

\begin{figure}
    \centering
    \includegraphics[width=.98\linewidth]{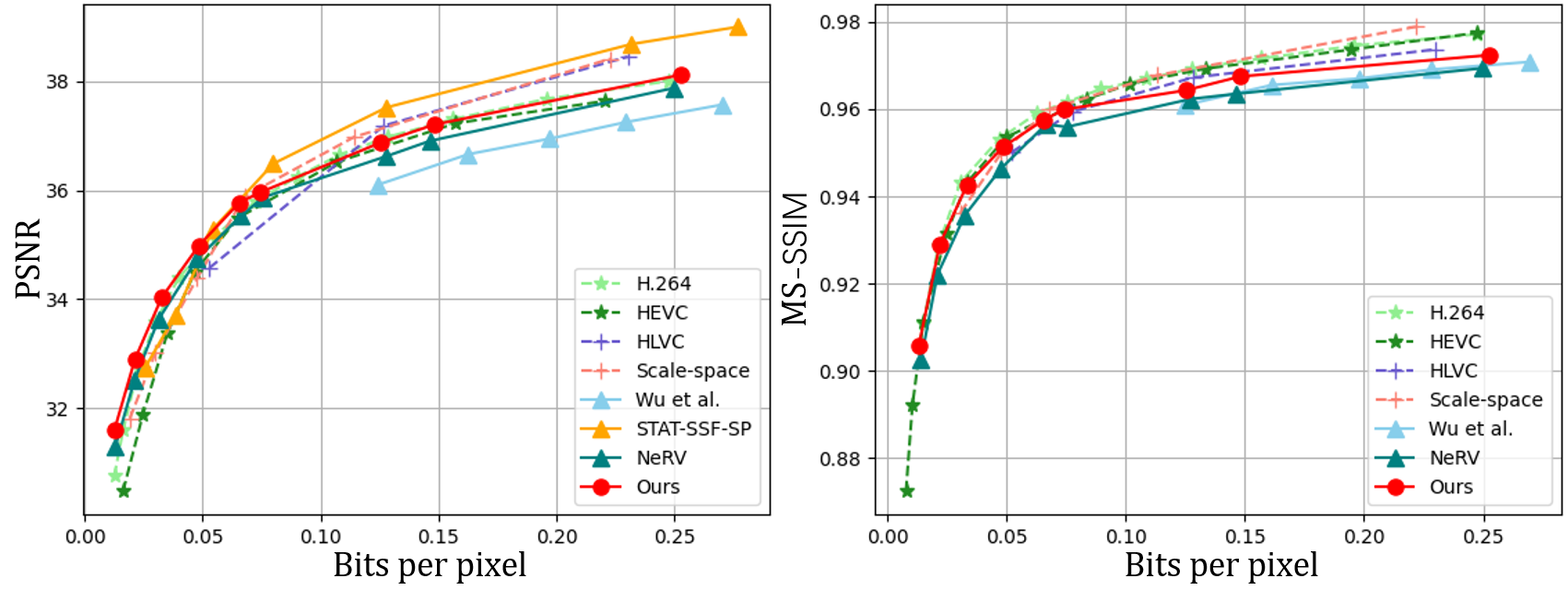}
    \vspace{-0.05in}
    \caption{PSNR and MS-SSIM \vs BPP on UVG dataset.}
    \label{fig:psnrssim}
    \vspace{-0.1in}
\end{figure}

\subsection{Objective Function}

For \system, we adopt a similar loss function as \cite{DBLP:conf/nips/ChenHWRLS21}, which combines L1 and SSIM loss for network optimization.
This function calculates the loss between the output patch and the ground-truth patch.
In order to reduce the difference between patches in the same frame, we add an additional total variation regularization $L_{tv}$. The final function is as follows:
\begin{equation}
    L = \frac{1}{T*N^2} \sum_{t=1}^{T}\sum_{i=1}^{N^2} (\alpha \left\lVert f_\theta(t,i) - v^t_{p_i}\right\rVert_1 + (1 - \alpha) (1 - \text{SSIM}(f_\theta(t,i), v^t_{p_i}))) + L_{tv} ,
    \label{equa:loss}
\end{equation} where $T$ is the frame number, $N^2$ is the number of patches in each frame, $f_\theta(t,i) \in  \mathbb{R}^{H/N\times W/N \times 3}$ is the \system prediction, $v^t_{p_i} \in  \mathbb{R}^{H/N\times W/N \times 3}$ is the ground-truth, $\alpha$ is hyper-parameter to balance the weight for each loss component.

\begin{figure}
    \centering
    \includegraphics[width=.98\linewidth]{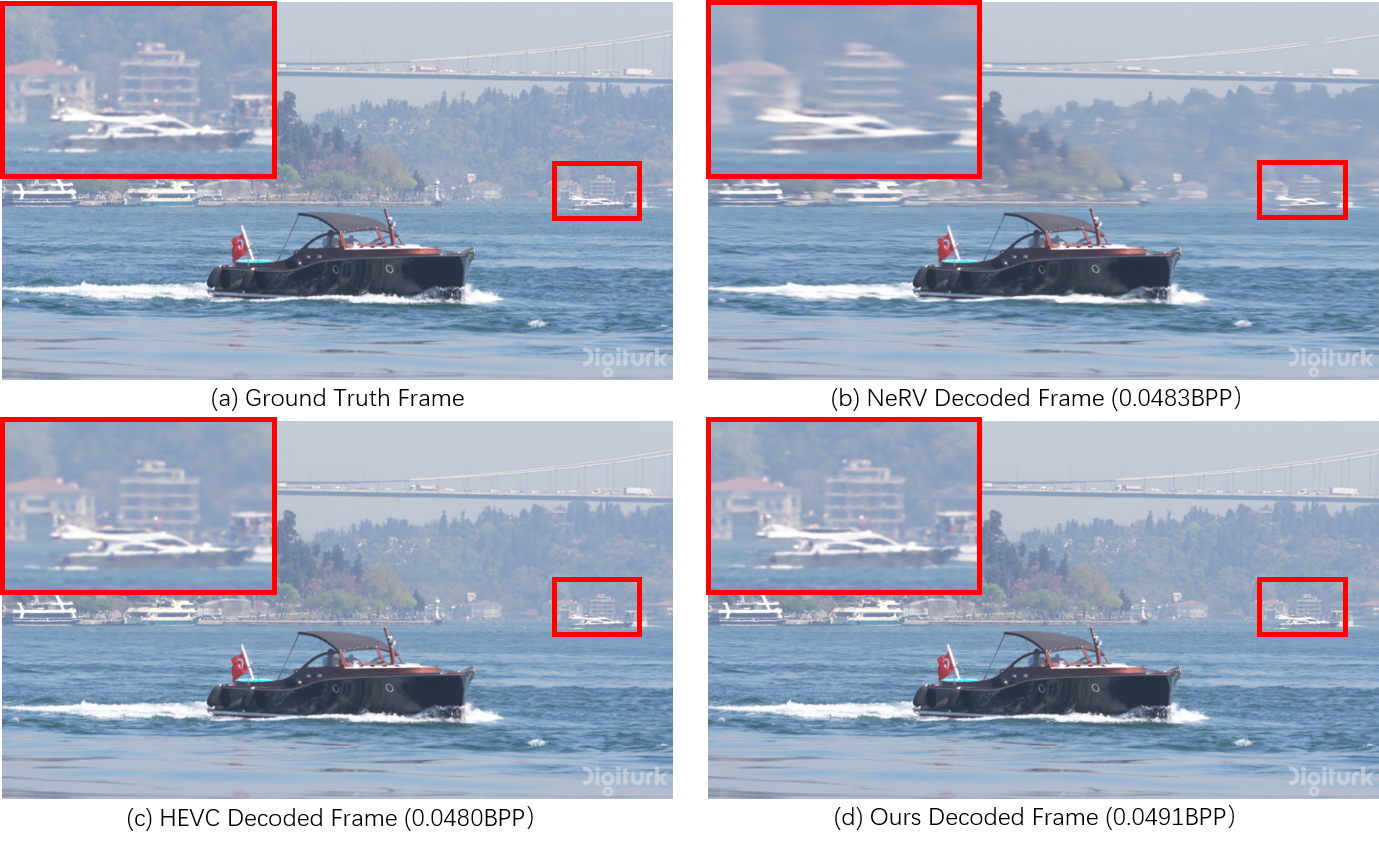}
    \vspace{-0.05in}
    \caption{Video compression visualization. At similar BPP, \system  reconstructs videos with better details.}
    \label{fig:Visualization for video compression.}
    \vspace{-0.1in}
\end{figure}

\section{Experiments}
\label{section4}
\vspace{-0.1in}
\subsection{Datasets and Implementation Details}
\vspace{-0.1in}
We use the ``Big Buck Bunny'' sequence, with 132 frames of $720\times1080$ resolution, as training data to conduct the experiments and compare the results with other INR methods. The video compression experiments are carried out on the widely used UVG~\cite{mercat2020uvg} dataset, which has 7 videos and 3900 frames with $1920\times 1080$ in total.
We use Adam \cite{DBLP:journals/corr/KingmaB14} optimizer to train the whole network. The learning rate is set to 5e-4. In the training process, we use cosine announcing learning rate schedule \cite{DBLP:conf/iclr/LoshchilovH17}, and the number of warm epochs is set to $30\%$ of all epochs. We trained 1200 epochs for experiments on ``Big Buck Bunny'', and 150 epochs on UVG experiments. There are five PSB blocks in our whole model, and the up-scale factors will be adjusted according to different patch split numbers. In the following experiments, the patch number is set to 16 unless otherwise denoted, and the up-scale factors are set to 5, 3, 2, 1, 1 for 1080p videos and 5,2,2,1,1 for 720p videos. For input embedding in Equation~\ref{equa:input-embed}, we use $b=1.25$ and $l=80$ as our default setting. For loss objective in Equation~\ref{equa:loss}, $\alpha$ is set to $0.7$. We only use one layer of MLP to get the mean and variance of later AdaIN layers, and its units are twice the number of these convolution layers' channels.

\subsection{Comparison with other INR methods}
\vspace{-0.1in}
We first compare our method with pixel-wise and image-wise INR methods. All models are trained on the ``Big Buck Bunny'' sequence for the same time.
SIREN~\cite{sitzmann2020implicit} and NeRF~\cite{mildenhall2020nerf} keep the original structure, use $sine$ activation functions and use positional embedding, respectively. For NeRV \cite{DBLP:conf/nips/ChenHWRLS21}, we also use its default settings. We obtain these models with different parameters by adjusting hidden dimension. 
We change the convolution filter width to build \system model of comparable sizes to the above models, named as \system-S, \system-M, and \system-L. The PSNR is served as the metric to evaluate reconstructed video quality. Table \ref{tab:compare} shows the comparison results. 
Compared with pixel-wise and image-wise methods, our patch-wise representation significantly improves the quality. Due to the addition of AdaIN and one more layer of MLP, the decoding speed will decrease a little, but remain on the same order as NeRV.

\begin{table}[t!]
    \begin{minipage}{.6\textwidth}
    \centering
        \caption{
Comparison with other INR methods.}
    \label{tab:compare}
    \centering
    \small
        \resizebox{.95\textwidth}{!}{
    \begin{tabular}{@{}l|ccccc@{}}
    \toprule
    Methods & Parameters  & PSNR $\uparrow$ & \makecell{Decoding \\ FPS $\uparrow$}\\
    \midrule
    SIREN~\cite{sitzmann2020implicit} & 3.2M  & 31.39 & 1.4 \\
    NeRF~\cite{mildenhall2020nerf}  & 3.2M  & 33.31 & 1.4 \\
    NeRV-S \cite{DBLP:conf/nips/ChenHWRLS21}  & 3.2M   & 34.12 & \textbf{35.48} \\
    \system-S (16) & 3.2M   & \textbf{35.82} & 32.43 \\
    \midrule
    SIREN~\cite{sitzmann2020implicit} & 6.4M  & 31.24 & 0.8 \\
    NeRF~\cite{mildenhall2020nerf} & 6.4M  & 35.18 & 0.8 \\
    NeRV-M~\cite{DBLP:conf/nips/ChenHWRLS21} & 6.3M  & 38.19& \textbf{34.66}\\
        \system-M (16) & 6.3M  & \textbf{40.29} & 31.87\\
    \midrule
    SIREN~\cite{sitzmann2020implicit} & 12.7M  & 25.12 & 0.4 \\
    NeRF~\cite{mildenhall2020nerf} & 12.7M  & 37.89 & 0.4 \\
    NeRV-L~\cite{DBLP:conf/nips/ChenHWRLS21} & 12.5M & 41.23 & \textbf{33.57} \\
        \system-L (16) & 12.5M & \textbf{43.06} & 30.98 \\
    \bottomrule
    \end{tabular}
    }
    \end{minipage}
    \hfill
    \begin{minipage}{.38\textwidth}
    \centering
    \small
    \caption{The ablation study of AdaIN. The parameters of these models are 3.2M, and they are all trained with 300 epochs. The introduction of AdaIN can significantly improve the effect of both image-wise and patch-wise methods.
    }
    \label{tab:style}
    \resizebox{.99\textwidth}{!}{
    \begin{tabular}{@{}l|c@{}}
    \toprule
    Methods & PSNR $\uparrow$ \\
    \midrule
    NeRV ~\cite{DBLP:conf/nips/ChenHWRLS21} & 32.04  \\
    NeRV-AdaIN  & 33.25\\
     \system w/o AdaIN (16) & 32.90 \\
    \midrule
    \system (16) & \textbf{34.34} \\
    \bottomrule
    \end{tabular}
    }
    \end{minipage}
    \vspace{-0.05in}
\end{table}

\begin{figure}
    \centering
    \includegraphics[width=.98\linewidth]{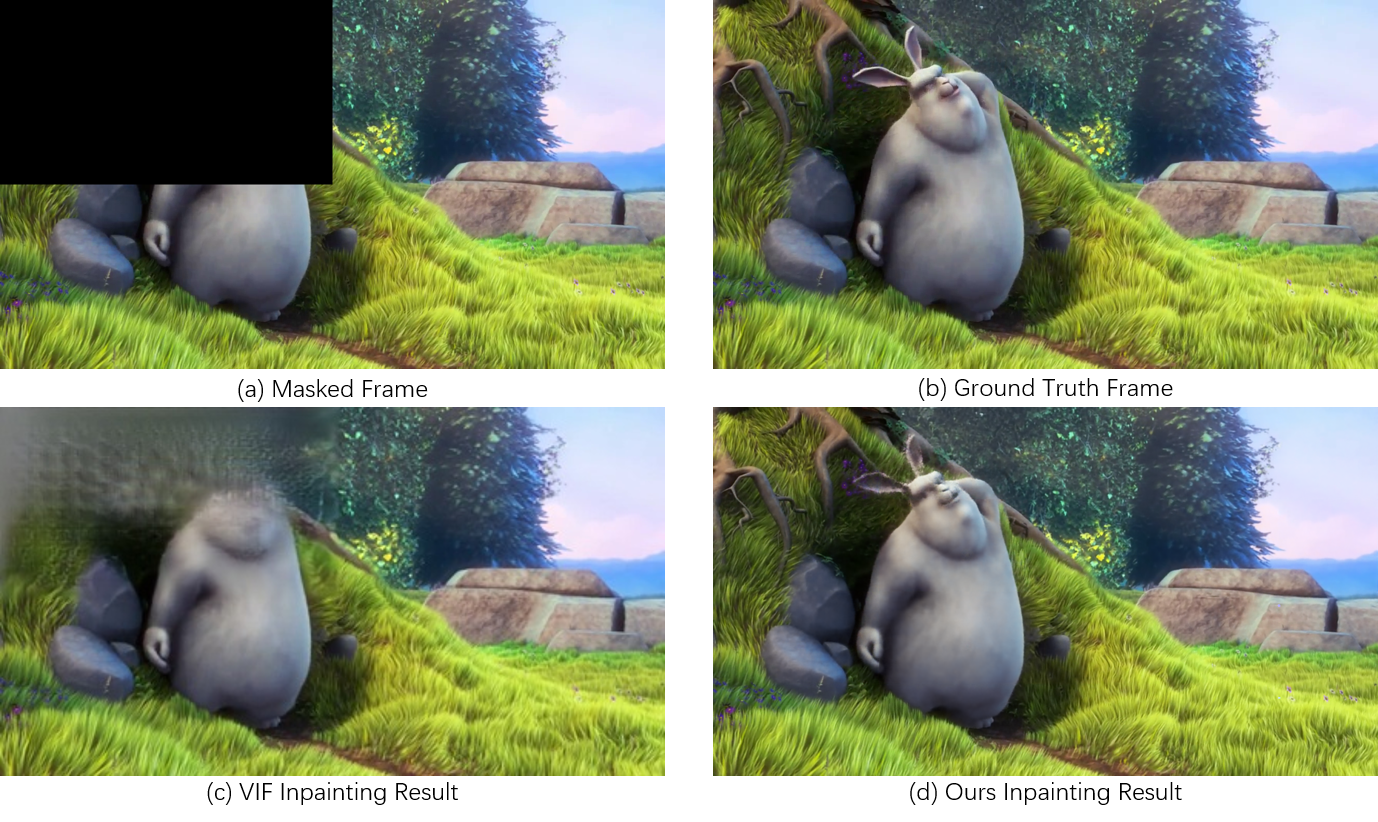}
    \vspace{-0.05in}
    \caption{Inpainting visualization. When a large area in the video is masked, our method can also get a good completion effect, while other learning-based method can only get blurry results.}
    \label{fig:inpainting}
    \vspace{-0.1in}
\end{figure}

\subsection{Video Compression}
\vspace{-0.1in}
Once the video fitting is completed, the purpose of video compression can be achieved through the model compression. For the fairness of comparison, we use the same model compression method as NeRV to achieve the purpose of video compression. The model compression process composes of three sequential steps: model pruning, weight quantization, and weight encoding. The model size reduction is realized by global unstructured pruning. When the weight is below a threshold, it will be set to $0$. The model quantization is carried out after training. Through the Equation used in NeRV, each parameter can be mapped to a `bit' length value. The  Huffman Coding~\cite{huffman1952method} scheme is also used to further compress the model size.
We then compare with state-of-the-arts methods on UVG dataset. Like the practice in NeRV, we concatenate 7 videos into one single video for training. Figure~\ref{fig:psnrssim} shows the rate-distortion curves. We compare  with H.264~\cite{wiegand2003overview}, HEVC~\cite{sullivan2012overview}, STAT-SSF-SP~\cite{yang2020hierarchical}, HLVC~\cite{yang2020learning}, Scale-space~\cite{agustsson2020scale}, and Wu \etal~\cite{wu2018video}. H.264 and HEVC are performed with \textit{medium} preset mode. Our method is better than image-wise method in all cases. When the BPP is small, our method even exceeds the traditional video compression technology and other learning-based video compression methods. Figure~\ref{fig:Visualization for video compression.} shows visualization for decoding frames. At a similar setting, \system reconstructs more accurate details.

\subsection{Video Inpainting}
\vspace{-0.1in}
Video inpainting is a task that aims at filling missing regions in video frames with plausible contents by fusing spatio-temporal information. Recent methods have designed complex models to solve this problem, such as Visual Transformers \cite{liu2021fuseformer} and Conv-LSTM \cite{kim2019deep}. While our method can achieve the goal of video inpainting by simply fitting the incomplete video. Specifically, given a masked video, the patches containing missing regions are not sampled during the training process. When the training is completed, we can input the coordinates of these missing regions to the network to obtain the corresponding image patch. We compare our results with the state-of-the-art transformer-based method ViF \cite{liu2021fuseformer}. As shown in Figure \ref{fig:inpainting}, in the case of large missing areas, ViF \cite{liu2021fuseformer} has difficulty generating meaningful content and can only fill in very blurry results. In contrast, by establishing an accurate mapping between frame patch and spatio-temporal coordinates, our method yields clear results that are almost indistinguishable from ground truth.






\begin{figure}
    \centering
    \includegraphics[width=.98\linewidth]{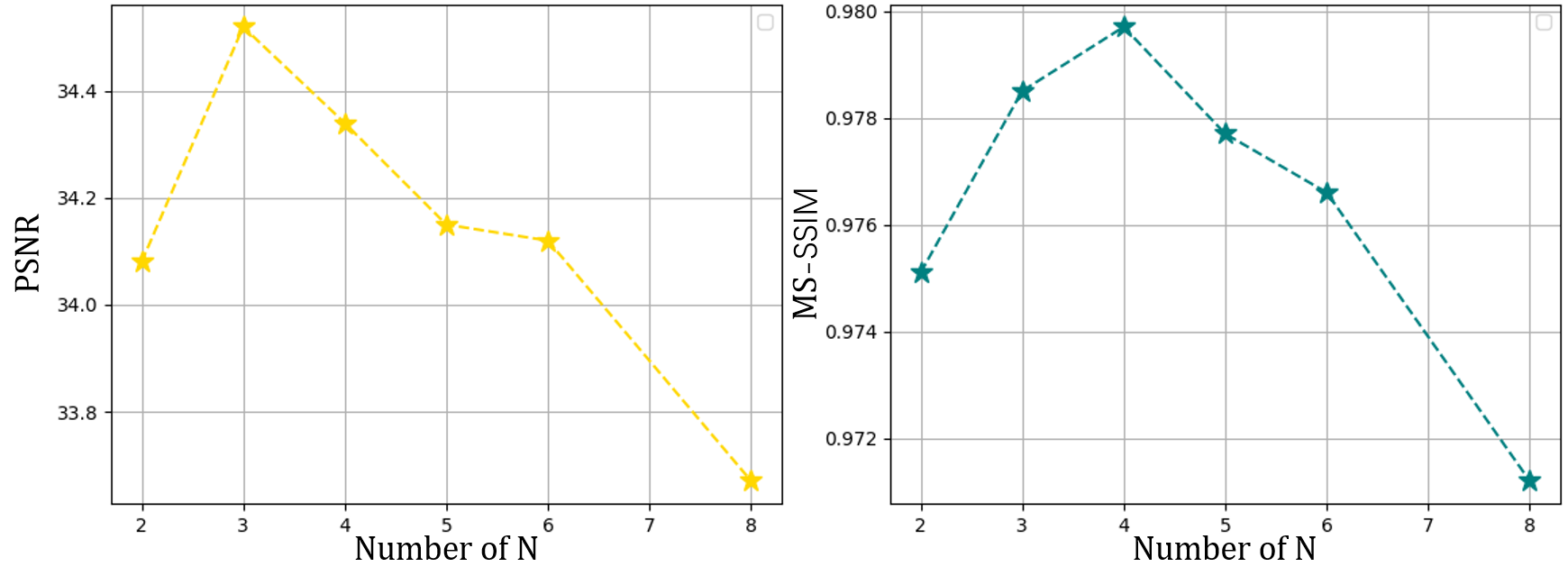}
    \vspace{-0.05in}
    \caption{PSNR and MS-SSIM \vs number of N.}
    \label{fig:patch}
    \vspace{-0.1in}
\end{figure}


\begin{figure}
    \centering
    \includegraphics[width=.98\linewidth]{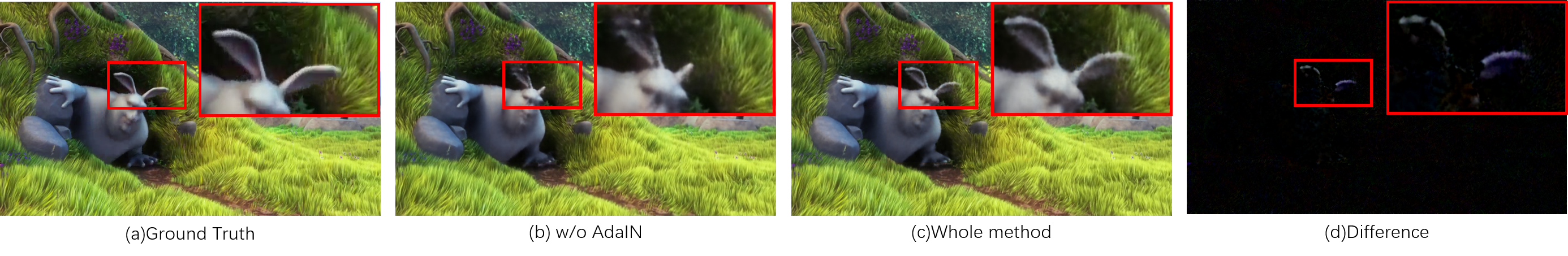}
    \vspace{-0.05in}
    \caption{The ablation study of AdaIN. Modulating features through AdaIN can restore more accurate details.}
    \label{fig:style}
    \vspace{-0.1in}
\end{figure}

\subsection{Ablation Study}
\vspace{-0.1in}
Two key components of our method are the patch-wise representation and stylized modulation by of features by AdaIN. In order to verify the effect of each part, we also conduct ablation studies on the ``Big Buck Bunny'' sequence. We first study the impact of the number of patches on the results. We design different upscale factors for different patch numbers, and change the filter width to get the same size model. As shown in the Figure \ref{fig:patch}, when the number of patches increases, the PSNR will decrease. This is because when the patch number increases, the method will become closer to pixel-wise, resulting in the decline of fitting efficiency. The appropriate patch number may vary between different videos, while too many patches will always degrade the quality and efficiency.

\begin{figure}
    \centering
    \includegraphics[width=.98\linewidth]{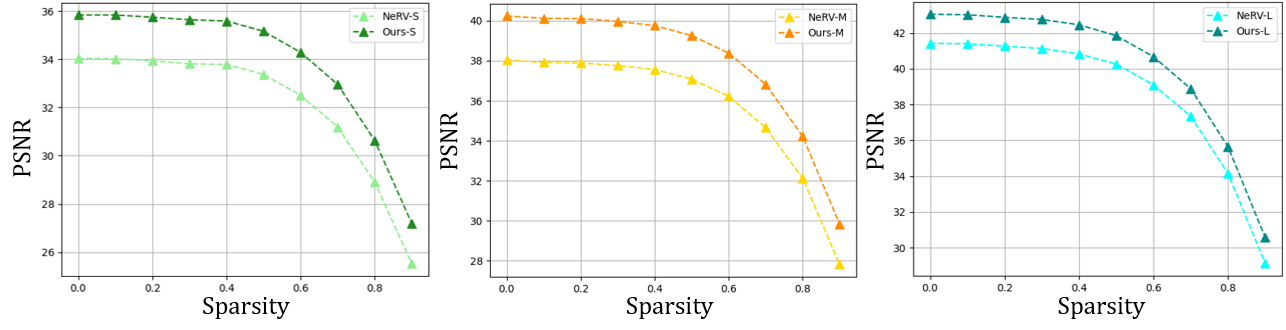}
    \vspace{-0.05in}
    \caption{Model \textbf{pruning}. Sparsity is the ratio of parameters pruned. The model of 40\% sparsity still reaches comparable performance with the original one.}
    \label{fig:bunny_prune}
    \vspace{-0.1in}
\end{figure}

\begin{figure}
    \centering
    \includegraphics[width=.98\linewidth]{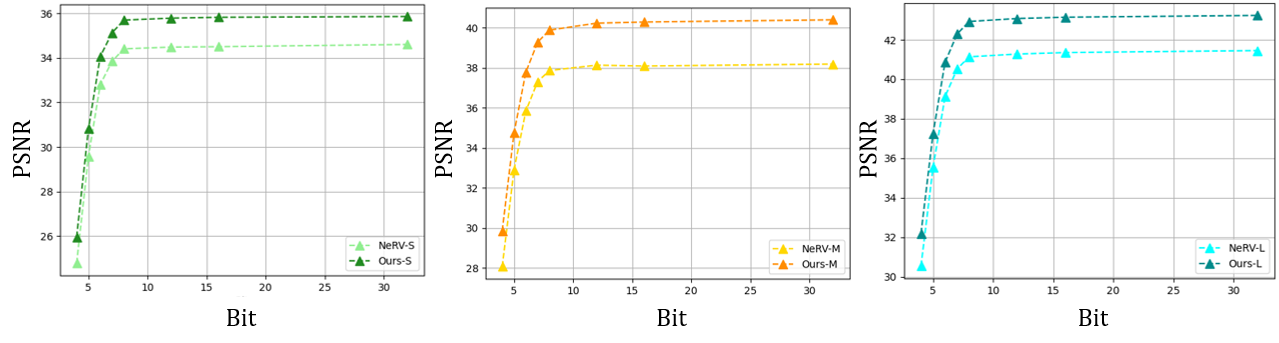}
    \vspace{-0.05in}
    \caption{Model \textbf{quantization}. Bit is the bit length used to represent parameter value. 8-bit model still remains the video quality compared to the full model.}
    \label{fig:bunny_quantize}
    \vspace{-0.1in}
\end{figure}

Then, in order to study the effect of AdaIN, we remove this layer from the architecture. 
Furthermore, we also introduce AdaIN into the framework of NeRV to modulate the features in the same way.
It can be seen from Table \ref{tab:style} that this practice can improve the quality of both patch-wise and image-wise methods. Figure \ref{fig:style} shows a comparison example. The picture obtained by \system on the right has more accurate and rich details. We also conduct ablation study on the model compression and compared the results with NeRV. Figure~\ref{fig:bunny_prune} and Figure~\ref{fig:bunny_quantize} shows the results of different pruning ratios and quantization step, respectively. Our method exceeds NeRV under various compression conditions. 



\section{Discussion}
\label{section5}
\vspace{-0.1in}
\textbf{Limitations and Future Work.} 
There are some limitations with the proposed \system. First of all, our patch-wise representation will increase the demand for video memory in the training process. Fortunately, even the common 1080Ti GPU can meet the training need of less than 64 patches for 1080p videos. In addition, we also train the whole network from scratch. In order to ensure the quality of video reconstruction, we still need a longer training time than the encoding time of traditional video compression methods. In the future work, we may introduce some meta-learning methods to improve the efficiency of network training, but this is not the goal of this work. Finally, the model compression method is also worthy of further exploration.





\textbf{Conclusion.}
In this work, we explore a more suitable implicit representation method for video. Inspired by the widespread local similarity in the real-world signals, we propose a more effective patch-wise INR method for video. We find that compared with the previous pixel-wise and image-wise methods, patch-wise representation method takes into account both efficiency and accuracy. Extensive experiments show that our method can be applied to several video-related applications, such as video compression and video inpainting. Patch-wise implicit representation may be an important means of video representation in future research. Moreover, considering the great potential of INR method, it may replace the traditional video representation in the future.

{\small
\bibliographystyle{plain}
\bibliography{egbib}
}

\end{document}